\documentclass[journal]{IEEEtran}

\usepackage{cite}
\usepackage{graphicx}
\usepackage{amsmath,amssymb,amsfonts}
\usepackage{algorithm}
\usepackage{algpseudocode}
\usepackage{mathtools}  
\usepackage{physics}    

\begin{document}

\title{Hybrid State Space-based Learning for Sequential Data Prediction with Joint Optimization}

\author{Mustafa~E.~Aydin, Arda~Fazla
        and~Suleyman~S.~Kozat,~\IEEEmembership{Senior~Member,~IEEE}
\thanks{This work is in part supported by Turkish Academy of Sciences Outstanding Researcher fund. 

The authors are with the Department of Electrical and Electronics Engineering, Bilkent University, 06800, Ankara, Turkey (e-mail: enesa@ee.bilkent.edu.tr; arda@ee.bilkent.edu.tr; kozat@ee.bilkent.edu.tr).}}


\maketitle

\begin{abstract}
We investigate nonlinear prediction/regression in an online setting and introduce a hybrid model that effectively mitigates, via a joint mechanism through a state space formulation, the need for domain-specific feature engineering issues of conventional nonlinear prediction models and achieves an efficient mix of nonlinear and linear components. In particular, we use recursive structures to extract features from raw sequential sequences and a traditional linear time series model to deal with the intricacies of the sequential data, e.g., seasonality, trends. The  state-of-the-art ensemble or hybrid models typically train the base models in a disjoint manner, which is not only time consuming but also sub-optimal due to the separation of modeling or independent training. In contrast, as the first time in the literature, we jointly optimize an enhanced recurrent neural network (LSTM) for automatic feature extraction from raw data and an ARMA-family time series model (SARIMAX) for effectively addressing peculiarities associated with time series data. We achieve this by introducing novel state space representations for the base models, which are then combined to provide a full state space representation of the hybrid or the ensemble. Hence, we are able to jointly optimize both models in a single pass via particle filtering, for which we also provide the update equations. The introduced architecture is generic so that one can use other recurrent architectures, e.g., GRUs, traditional time series-specific models, e.g., ETS or other optimization methods, e.g., EKF, UKF. Due to such novel combination and joint optimization, we demonstrate significant improvements in widely publicized real life competition datasets. We also openly share our code for further research and replicability of our results.
\end{abstract}



\section{Introduction}\label{sec:introduction}
\subsection{Background}\label{sec:background}
\IEEEPARstart{W}{e} study nonlinear prediction/regression in a sequential/online setting where we receive a data sequence associated with a target sequence and predict this sequence's next samples. This problem is extensively studied in the machine learning literature as it has many practical applications, e.g., in predicting electricity demand \cite{fore_elec}, weather conditions \cite{fore_weather} and medical records \cite{fore_medic}. Generally, this problem is studied as two disjoint sub-problems where first explanatory variables are extracted possibly by a domain expert and afterwards a regressor is trained over these features for the ultimate prediction. With the introduction of neural networks (especially recurrent ones \cite{lstm_itself, lstm_odssey}), however, there have been attempts to automate this procedure by unifying the objectives into a single machinery and feature extraction as a joint task. Recurrent neural networks (RNN), such as a long short-term memory (LSTM) network, are highly complex models used in demystifying nonlinear relationships between the input and output pairs. This can easily lead to overfitting; and in particular, for our context of time series prediction, they tend to ``memorize'' the last seen sample and output that as the prediction, i.e., behave similar to a naive model \cite{diger_paper}. Furthermore, these complex networks tend to shift their focus to apparent seasonalities and linear trends inherent to time series data, as these patterns are rather easy for the network to notice and learn from. However, this leads to a ``distraction'' of the network from figuring out the more vital nonlinear relationships in the data as shown in our simulations, where a single RNN readily tends to converge to a (seasonal) naive (possibly with a drift) model on time series prediction, i.e., becomes a simple model due to ``focusing'' too much on specific characteristics inherent to a time series such as trend and seasonality. 

On the other hand, there exist ``traditional'' time series models, such as auto-regressive moving average (ARMA) (or its extension seasonal auto-regressive integrated moving average with exogenous regressors (SARIMAX)) and exponential smoothing (ETS) \cite{hyndman2018forecasting}, that are linear and rather simple models compared to neural networks. These models have been designed with time series properties in consideration. Therefore, they heavily possess problem-specific parts in their formulation, e.g., a SARIMAX model deals with seasonality directly (among other things) and an ETS model accounts for trend and seasonality components of a sequence right in its formulation. However, these models have too strong assumptions about the data such as linearity and stationarity that disallow them to capture nonlinear patterns that most real life data tend to possess \cite{nonlinear_life}.

Here, we introduce a unified time series-specific model and a recurrent neural network to achieve a ``separation of duties'' in prediction, i.e., the RNN is tasked for automatic feature extraction and revealing the nonlinear relations while a time series-specific model, e.g., SARIMAX, is employed for trend and seasonality detection in a linear manner. Moreover, the exogenous variables, i.e., the side information that help in prediction, undergo numerous nonlinear transformations in an RNN structure to produce finally the prediction output. Whereas in a less complex SARIMAX model, there is a direct linear link between the side information and the desired signal. Therefore, we combine/ensemble an RNN with a SARIMAX model side by side to bring their prominent features seamlessly for effective prediction while enjoying joint optimization of the parameters of the entire model. In particular, we merge the two models side-by-side by putting them into a joint nonlinear state space formulation and jointly optimize the whole architecture using particle filtering. In order to achieve online learning, we introduce novel state space formulations, which are merged, allowing for a single-pass solution for optimal parameter selection as the first time in the literature. We highlight that our framework is generic, i.e., one can use other recurrent network architectures for the nonlinear part (such as simple RNNs and GRUs), traditional time series models (such as ETS) for the linear part and other optimization models (such as Extended Kalman Filtering (EKF), Unscented Kalman Filtering (UKF)) as demonstrated in the ablation studies in our experiments.

\subsection{Related Work}\label{sec:relatedwork}
Neural networks have been used for time series prediction for decades now \cite{nn_ts_1, nn_ts_2}. With the introduction of powerful recurrent neural networks \cite{lstm_itself}, focus shifted towards RNN-based solutions for time series tasks. Even though they are not particularly designed for the task, their sequential data processing power easily allowed them to be popular in the area \cite{more_recent_nn_ts_1, more_recent_nn_ts_2}. However, despite their ability to work with raw features, neural networks need a lot of domain-specific preprocessings and feature engineering for sequential data prediction tasks \cite{ts_specific_nn_1, ts_specific_nn_2}; otherwise they are known to converge to a naive predictor, i.e., a simple model outputting the last observation as the final prediction \cite{lstm_naive, diger_paper}. On the other hand, traditional time series models, such as ARMA and ETS, date back even more than neural networks. They have generally fewer parameters to estimate, are easier to interpret due to their naturally intuitive modeling of the underlying sequence and are robust to overfitting and amenable to automated procedures for hyperparameter selection \cite{the_forecasting_paper}. Since they are tailored specifically for time series regression, they are still employed in contemporary forecasting competitions \cite{the_forecasting_paper}. Nevertheless, they have highly strong assumptions about the data such as stationarity and linearity, which prevent them from capturing nonlinear patterns that most real life data tend to possess \cite{nonlinear_life}. Therefore, both the (nonlinear) complex recurrent networks and the (linear) traditional models have their drawbacks when it comes to time series prediction.

To overcome the individual weaknesses of each model families, there have been several works on combining/ensembling linear and nonlinear models. One of the most common strategies is to build a separate machine learning model, often referred to as the ``meta-learner'' (or ensemble learner) \cite{meta_learner_generic}, which learns to combine the forecasts of multiple models. The key problem with the given approach is that the linear and the nonlinear models in the architecture are disjointly optimized, as the forecasts of the models are provided to the ensemble learner independent of the model-building phase. Consequently, the linear and nonlinear models do not mutually benefit each other or the ensemble learner during the training process, which can lead to sub-optimal results.

Specifically, several works in the machine learning literature have focused on combining/ensembling an RNN with an ARMA model. In \cite{Nazaripouya}, authors successively combine an ARMA model with a recurrent neural network after preprocessing with wavelet transformations for solar power prediction. However, the combination is done in a \emph{disjoint} manner, i.e., only after an ARMA model is fit on the data, is the RNN employed on the residuals. This disjoint approach not only increases the computational time but also retains it from enjoying the benefits of a joint optimization of an integrated model \cite{end_to_end_paper_1,end_to_end_paper_2}. Moreover, they focus on solar power prediction, which renders the overall model domain specific. In \cite{Li_deng}, authors alter the recurrent relation of a vanilla RNN for its hidden state via an inspiration from ARMA models by incorporating the effects of lagged input for the state transition. As their experiments show, however, this does not cause a dramatic increase in performance compared to the vanilla RNN structure, which is possibly because the side information sequence, which does not directly affect the final output in their framework, can be highly auto-correlated and recent lagged values do not carry as much information. Furthermore, they focus on sequence classification tasks and do not address time series prediction. In \cite{Yongjin_Jeong}, inspiration is this time taken from RNN models towards an ARMA-based model. Authors subject auto regressive (AR) and moving average (MA) components to sigmoid nonlinearities borrowed from RNNs. Nonetheless, this does not bring the full capabilities of an enhanced RNN, e.g., the gating mechanisms for efficient information flow is missing. 

Overall, the contemporary literature addresses the unification of models either in a disjoint manner or via borrowing ideas from one model family to another to implement partial integration. Unlike previous studies, our model as the first time in the literature achieves joint optimization of a nonlinear and recurrent sequential feature extractor (e.g., LSTM) and a linear time-series specific model (e.g., SARIMAX) for online learning, where both models are unified through a state space formulation thanks to our novel state space structures. Hence, as shown in our simulations, nonlinear and linear parts in a sequence are fully learned, providing significant improvement in performance over the well-known real life competition datasets compared to the state-of-the-art machine learning models. 

\subsection{Contributions}\label{sec:contributions}
We list our main contributions as follows:
\begin{enumerate}
    \item We introduce a hybrid architecture composed of an LSTM network and a SARIMAX model for sequential data prediction that is trainable in a joint manner with particle filtering. To the best of our knowledge, this model is the first in the literature that is armed with joint optimization for a nonlinear sequential feature extractor and a linear traditional time series regressor.
    \item We introduce novel state space formulations that allow for a single-pass solution for parameter estimation, which eases online learning and concatenation of state spaces with other models.
    \item The introduced architecture is generic enough that the nonlinear part can be replaced with, for example, another RNN variant (e.g., gated recurrent unit (GRU) \cite{gru_paper}) or a temporal convolutional network (TCN) \cite{tcn_paper}. Similarly, the linear part can also be another time series regression model (e.g., ETS \cite{ets}) and joint learning can be accomplished by other state space optimization methods, where we use the particle filtering due to its well-known superior performance in state space optimization \cite{PF_theory_1}.
    \item Through an extensive set of experiments, we show the effectiveness and efficacy of the introduced model over the well-known competition datasets.
    \item We publicly share our code\footnote{https://github.com/mustafaaydn/lstm-sx} to facilitate further research and ensure reproducibility of our results.
\end{enumerate}

\subsection{Organization}\label{sec:organization}
The rest of the paper is organized as follows. In Section \ref{sec:preliminaries}, we introduce the nonlinear prediction problem, describe the components of the hybrid/ensemble architecture, i.e., the LSTM neural network and the SARIMAX model. We then present the state space formulations of the individual models and the introduced joint model along with the update equations for the particle filter that solves the prediction problem in Section \ref{sec:theproposedmodel}. We demonstrate the performance of our model through several experiments over the well-known real life datasets in Section \ref{sec:experiments}. Finally, Section \ref{sec:conclusion} concludes the paper.

\section{Preliminaries}\label{sec:preliminaries}
\subsection{Problem Statement}\label{sec:problemstatement}
In this paper, all vectors are column vectors, represented by boldface lowercase letters and all the entries are real numbers. Matrices are denoted by boldface uppercase letters. $\boldsymbol{x}_{i:j}$ represents a slice of the vector $\boldsymbol{x}$ from its $i$\textsuperscript{th} component to $j$\textsuperscript{th} component, both inclusive. $\boldsymbol{x}^T$ represents the ordinary transpose of $\boldsymbol{x}$. ${X}_{i, j}$ represents the entry at the $i\textsuperscript{th}$ row and the $j\textsuperscript{th}$ column of the matrix $\boldsymbol{X}$. 

We study online regression of sequential data. We observe a sequence $\{y_t\}$ along with a side information sequence (or feature vectors) $\{\boldsymbol{s}_t\}$. At each time $t$, given the past information $\{y_k\}$, $\{\boldsymbol{s}_k\}$ for $k \leq t$, we produce the prediction $\hat{y}_{t+1}$. In this fully online setting, our goal is to find the relationship
\begin{equation*}
    \hat{y}_{t+1} = f\big(\{\,\ldots, y_{t-1}, y_t\}, \{\,\ldots, \boldsymbol{s}_{t-1}, \boldsymbol{s}_{t}\}\big),
\end{equation*}
where $f$ is an unknown function, which models $\hat{y}_{t+1}$. In our framework, $\hat{y}_{t+1}$ is formed as a sum or ensemble of two components: prediction from the linear and the nonlinear models, i.e.,
\begin{equation}\label{eq:nonlinear_and_linear_comb}
    \hat{y}_{t+1} = \hat{y}_{t+1}^{(n)} + \hat{y}_{t+1}^{(l)},
\end{equation}
where $\hat{y}_{t+1}^{(n)}$ and $\hat{y}_{t+1}^{(l)}$ denote the nonlinear and linear parts, e.g., coming from an LSTM network and a SARIMAX model, respectively. We emphasize that we do not use any weights in combination since, as will be demonstrated, they will be handled by our joint optimization framework. 

We note that even though these predictions are coming from different formulations, which are extensively used in the machine learning literature, our framework encapsulates them both in a unified state space representation and optimizes the underlying models jointly. 

Once $y_{t+1}$ is revealed, we suffer the loss
\begin{equation*}
    L_{t+1} = \ell(y_{t+1}, \hat{y}_{t+1}),
\end{equation*}
where $\ell$ can be, for example, the squared error loss, i.e., $\ell(y_{t+1}, \hat{y}_{t+1}) = (y_{t+1} - \hat{y}_{t+1})^2$. As an example, in weather nowcasting, $y_t$ can represent the temperature at this hour and the next hour is predicted using the hourly measurements of the past. Side information, i.e., $\boldsymbol{s}_t$, could include the wind, humidity levels or features derived from the past information, e.g., average of last 7 hours etc. The model may adapt itself whenever new observation of the next hour becomes available, i.e., we work in an online manner constantly adapting to the changes in time.

We use RNNs as the nonlinear part of the framework since RNNs and its variants are widely used to model sequential data \cite{rnn_ts_1, rnn_ts_2}. An RNN is described by the recursive equation \cite{rnn_itself}
\begin{equation*}
    \boldsymbol{h}_t = g(\boldsymbol{W}_{ih} \boldsymbol{x}_t + \boldsymbol{W}_{hh} \boldsymbol{h}_{t-1}),
\end{equation*}
where $\boldsymbol{x}_t$ is the input vector and $\boldsymbol{h}_t$  is the state vector at time instant $t$. $\boldsymbol{W}_{ih}$ and $\boldsymbol{W}_{hh}$ are the input-to-hidden and hidden-to-hidden weight matrices (augmented to include biases), respectively; $g$ is a nonlinear function that applies element wise. As an example, for the prediction task, one can use $\boldsymbol{s}_t$ directly as the input vector, i.e., $\boldsymbol{x}_t = \boldsymbol{s}_t$, or some combination of the past information $\{y_t, \ldots, y_{t-r+1}\}$ with a window size $r$ along with $\boldsymbol{s}_t$ to construct the input vector $\boldsymbol{x}_t$. The final output of the RNN can be produced by passing $\boldsymbol{h}_t$ through a dense layer on top:
\begin{equation}\label{eq:lstm_dense}
    \hat{y}_{t+1} = \boldsymbol{w}_t^T \boldsymbol{h}_t,
\end{equation}
where $\boldsymbol{w}_t, \boldsymbol{h}_t \in \mathbb{R}^m$. Note that because of the generic $\boldsymbol{w}_t$ here, we do not need to include combination weights in \eqref{eq:nonlinear_and_linear_comb}. 

To effectively deal with the temporal dependency in sequential data, we use a specific kind of RNN, the LSTM neural networks \cite{lstm_itself}, which has shown significant results in a wide range of highly publicized competitions \cite{m4_comp, m5_comp}. Here we use the most widely used variant with the forget gates and without peephole connections \cite{lstm_forget}. The forward pass equations in one cell are described as
\begin{equation}\label{eq:lstmforward}
\begin{aligned}
    \boldsymbol{f}_t &= \sigma(\boldsymbol{W}_f \,\{\boldsymbol h_{t-1}, \boldsymbol x_t\} + \boldsymbol b_f)\\
    \boldsymbol{i}_t &= \sigma(\boldsymbol{W}_i \,\{\boldsymbol h_{t-1}, \boldsymbol x_t\} + \boldsymbol b_i)\\
    \tilde{\boldsymbol c}_t &= \tanh(\boldsymbol{W}_c \,\{\boldsymbol h_{t-1}, \boldsymbol x_t\} + \boldsymbol b_c)\\
    \boldsymbol{o}_t &= \sigma(\boldsymbol{W}_o \,\{\boldsymbol h_{t-1}, \boldsymbol x_t\} + \boldsymbol b_o)\\
    \boldsymbol{c}_t &= \boldsymbol{f}_t \odot \boldsymbol{c}_{t-1} + \boldsymbol{i}_t \odot \tilde{\boldsymbol c}_t\\
    \boldsymbol h_{t} &= \boldsymbol o_{t} \odot \tanh(\boldsymbol c_t),
\end{aligned}
\end{equation}
where $\boldsymbol c_t$ is the state vector, $\boldsymbol x_t$ is the input to the cell, $\boldsymbol h_{t-1}$ is the hidden state from the previous cell and $\{\boldsymbol h_{t-1}, \boldsymbol x_t\}$ denotes the vertical concatenation of $\boldsymbol x_t$ and $\boldsymbol h_{t-1}$. $\boldsymbol f_t, \boldsymbol i_t, \tilde{\boldsymbol c}_t, \boldsymbol o_t$ are the forget gate, input gate, block input and output gate vectors, respectively, whose stacked weight matrices are $\boldsymbol W_f, \boldsymbol W_i, \boldsymbol W_c$ and $\boldsymbol W_o$. The corresponding biases are represented by $\boldsymbol b_f, \boldsymbol b_i, \boldsymbol b_c$ and $\boldsymbol b_o$. $\sigma$ denotes the logistic sigmoid function, i.e, $\sigma(x) = 1 / (1 + \exp(-x))$ and $\tanh$ is the hyperbolic tangent function, i.e, $\tanh(x) = 2\sigma(2x) - 1$; both are nonlinear activation functions that apply point-wise. $\odot$ is the Hadamard (element-wise) product operator.
By allowing a constant error flow through the cell state in backpropagation, LSTMs significantly prevent vanishing gradients, and nonlinear gated relations described by \eqref{eq:lstmforward} manage the information flow effectively to handle long term dependencies \cite{lstm_odssey}. The final output of \eqref{eq:lstmforward} is produced as in \eqref{eq:lstm_dense}. 

As for the linear part of the framework, we employ the SARIMAX model \cite{box_jenkins_arma,some_modern_sarimax}, which is a state-of-the-art linear model used in time series prediction \cite{the_forecasting_paper}. Tailored specifically for time series modeling, SARIMAX builds on the conventional ARMA architecture \cite{some_signal_processor_arma} via considering the difference(s) of the target signal where seasonal effects and side information are also incorporated. The ARMA model of order $(p, q)$ on $y_t$ corresponds to
\begin{equation}\label{eq:arma_eqn}
    y_t = \sum_{i=1}^p \phi_i y_{t - i} + \sum_{j=1}^q \theta_j \varepsilon_{t-j} + \varepsilon_t,
\end{equation}
where $\varepsilon_t$ is an independent and identically distributed (i.i.d.) noise sequence, which can sampled from any distribution such as Gaussian, and $\phi_i$ and $\theta_j$ are the parameters to estimate for the auto-regressive and the moving average parts, respectively. To simplify the expression, we use the ``lag notation'', with which \eqref{eq:arma_eqn} becomes
\begin{equation*}
    \phi_p(B)\,y_t = \theta_q(B)\,\varepsilon_t,
\end{equation*}
where the lag polynomials are defined as $\phi_p(B) := (1 - \phi_1B - \ldots -\phi_pB^p)$ and $\theta_q(B) := (1 + \theta_1B + \ldots + \theta_qB^q)$ where $B$ is the backshift (lag) operator defined as $Bx_t := x_{t-1}$ for any sequence $x_t$ such that the recurrence $B^kx_t = B(B^{k-1}x_t)$ holds for all integers $k \ge 2$. The SARIMAX model of order $(p, d, q)$ and seasonal order $(P, D, Q)$ with seasonality $m$ is then expressed as
\begin{equation}\label{eq:sarimax_eq}
\begin{aligned}
    \tilde{y}_t &= \Delta_d(B)\,\Delta_D(B^m)\,y_t\\
    \phi_p(B)\,\Phi_P(B^m)\,\tilde{y}_t &= \theta_q(B)\,\Theta_Q(B^m)\,\varepsilon_t + \boldsymbol{\beta}_t^T \boldsymbol{x}_t,
\end{aligned}
\end{equation}
where $\Delta_h := (1 - B - \ldots - B^h)$ is the difference operator, $\{\tilde{y}_t\}$ is the differenced sequence, $\boldsymbol{x}_t$ (or $\boldsymbol{s}_t$) is the side information vector, i.e., exogenous context-based data to help in prediction, and $\boldsymbol{\beta}_t$ is the coefficient vector associated with it.

In the following section, we first the present state space models for both LSTM and SARIMAX architecture. We then introduce a joint state space model to accommodate both LSTM and SARIMAX models together and optimize it in an end-to-end manner. Building on top of that, we introduce a generic particle filter-based solution for the sequential prediction problem at hand.

\section{The Introduced Hybrid/Ensemble Model}\label{sec:theproposedmodel}
We next introduce a hybrid model which aims to boost the performance of RNNs in time series prediction using a time series-specific linear model as auxiliary while enjoying joint optimization.

RNNs are generally trained with backpropagation through stochastic gradient descent (SGD) \cite{lstm_odssey}. Time series specific models such as SARIMAX (or ETS), on the other hand, are usually trained with Bayesian estimation methods with iterative optimization \cite{box_jenkins_arma}, e.g., maximum likelihood estimation (MLE) where no prior information on parameters is assumed, after they are put in a state space form. A gradient-based solution is rendered ineffective for these type of models mainly because they include ``unobserved'' components, i.e., the current and past errors (shocks) used by the model, for which the gradient calculations are not possible as they are unknown. Even though there exist methods to approximate these current and lagged errors, e.g., Hannah and Rissanen's \cite{Hannah_Rissanen} technique of replacing the moving average $\text{MA}(q)$, i.e., the error related parts of a SARIMAX model with a sufficiently large auto regressive $\text{AR}(p)$ model, these are not suitable for online prediction as they require continuous re-fitting in a sliding window manner to keep track of the unobserved components. Therefore, to merge or ensemble an RNN and a SARIMAX model, we introduce a nonlinear joint state space model which can be efficiently trained with, e.g., particle filtering.

\textit{\textbf{Remark.}} We emphasize that after we introduce the joint state-space model one can use any other estimator to train the model such as EKF, UKF. We use the particle filtering approach since it is shown to give the best results in training the state space models \cite{some_pf_super_good_1, some_pf_super_good_2}.

Nevertheless, the contemporary state space formulations for SARIMAX modeling, e.g., Harvey's \cite{ss_harvey} or Hamilton's \cite{ss_hamilton} do not put the parameters to be estimated directly into the state vector but instead mix it with the unobserved components for a more compact state representation. This, however, renders single-pass solutions, e.g., with a Kalman filter \cite{KF}, unsuitable for parameter estimation since the matrices in the formulation of the state spaces cannot be built out of unknown components. Then, instead of a single pass estimation, an iterative optimization based on MLE is used, which requires many passes over the time series that is not possible for online prediction. Therefore, we introduce a novel state space formulation for SARIMAX that puts the parameters into the state vector and excludes the unobserved components. This not only allows for a single-pass solution suitable for online prediction but also makes it possible to concatenate state space representations with that of an RNN for a joint optimization.

\subsection{State Space Formulations}
Here, we first present the state space formulations of the base models. Then, we join them for the hybrid architecture and introduce a particle filter-based solution for the prediction problem.
\subsubsection{LSTM}
As presented in \eqref{eq:lstmforward}, an LSTM network has two inherent state variables: the cell state $\boldsymbol{c}_t$ and the hidden state $\boldsymbol{h}_t$. Their (nonlinear) evolution will be captured by the state transition equations as
\begin{align}\label{eq:lstm_internals}
    \boldsymbol{c}_t &= \gamma\,(\boldsymbol{x}_t, \boldsymbol{h}_{t-1}, \boldsymbol{c}_{t-1}) + \boldsymbol{u}_t\\
    \boldsymbol{h}_t &= \tau\,(\boldsymbol{x}_t, \boldsymbol{h}_{t-1}, \boldsymbol{c}_{t}) + \boldsymbol{v}_t,
\end{align}
where $\gamma$ and $\tau$ represent the nonlinear relations depicted in \eqref{eq:lstmforward}, and $\boldsymbol{u}_t$ and $\boldsymbol{v}_t$ are each i.i.d. noise sequences that can sampled from any distribution such as Gaussian. The noise components are additionally integrated into the standard LSTM equations, as they are essential for estimating the parameters of the LSTM network from its state space representation. Note that we not only estimate the internal states of an LSTM cell but also the parameters, i.e., the weights and biases in \eqref{eq:lstm_dense} and \eqref{eq:lstmforward}. For brevity, we collect all of these coefficients in a large parameter vector $\boldsymbol{\theta}_t \in \mathbb{R}^{n_\theta}$ which includes flattened $\boldsymbol{W}_*$ matrices and $\boldsymbol{b}_*$ vectors for $* = \{f, i, c, o\}$ and the final regressor vector $\boldsymbol{w}_t$. Assuming a state size of $k$ and input dimension of $l$, i.e., $\boldsymbol{h}_t \in \mathbb{R}^k$ and $\boldsymbol{x}_t \in \mathbb{R}^l$, we have $n_\theta = 4\,(k(k+l) + k) + k$. In order to put these parameters into a state equation, we introduce a time-varying system where the parameters are allowed to evolve with random noise in state transition for online learning. This also allows modeling a non-stationary process in general as a byproduct. Given this, and letting $\boldsymbol{s}_{LSTM}$ be the encompassing state vector covering the internal state as well as the weights and biases of the network, we have the overall state transition equation for an LSTM cell as
\begin{equation}\label{eq:lstm_state_transition}
    \boldsymbol{s}_{t,LSTM} \coloneqq \begin{bmatrix}
                             \boldsymbol{c}_t\\
                             \boldsymbol{h}_t\\
                             \boldsymbol{\theta}_t
                            \end{bmatrix} = \begin{bmatrix}
                             \gamma\,(\boldsymbol{x}_t, \boldsymbol{h}_{t-1}, \boldsymbol{c}_{t-1})\\
                             \tau\,(\boldsymbol{x}_t, \boldsymbol{h}_{t-1}, \boldsymbol{c}_{t})\\
                             \boldsymbol{\theta}_{t-1}
                            \end{bmatrix} +
                            \begin{bmatrix}
                             \boldsymbol{u}_t\\
                             \boldsymbol{v}_t\\
                             \boldsymbol{\epsilon}_{t}
                            \end{bmatrix},
\end{equation}
where $\boldsymbol{\epsilon}_{t}$ is the i.i.d. noise vector driving the parameters of the LSTM, which can be sampled from any distribution such as Gaussian. Lastly, the measurement equation is given as
\begin{equation}\label{eq:lstm_measurement}
        {y}_{t,LSTM} = \boldsymbol{w}_t^T \boldsymbol{h}_t + \tilde{\varepsilon_t},
\end{equation}
where $\tilde{\varepsilon_t}$ is a scalar noise, which can sampled from any distribution such as Gaussian, and is assumed to be independent of $\boldsymbol{u}_t$, $\boldsymbol{v}_t$ and $\boldsymbol{\epsilon}_t$ of the state transition. Hence, in this way, we represent the LSTM network equations in \eqref{eq:lstmforward} with a complete state space formulation.

\subsubsection{SARIMAX}
We first start with the well-known state space formulation of an $\text{ARMA}(p, q)$ model \cite{box_jenkins_arma}, where $p$ and $q$ are the order of auto-regressive and moving average parts, respectively, and then derive the representation for a $\text{SARIMAX}(p, d, q)(P, D, Q, m)$ model where $d$ and $D$ are the order of ordinary and seasonal differences to the sequence, $P$ and $Q$ are the seasonal counterparts of $p$ and $q$, and $m$ represents the period of seasonality of the sequence, if any, e.g., 24 for hourly data. Unlike the current state space approaches for the ARMA model, we (only) put parameters to be estimated into the state vector. From \eqref{eq:arma_eqn}, we introduce time variability to the AR and MA parameters as
\begin{equation}\label{eq:arma_eqn_tv}
    y_t = \sum_{i=1}^p \phi_{t,i}\,y_{t - i} + \sum_{j=1}^q \theta_{t,j}\,\varepsilon_{t-j} + \varepsilon_t,
\end{equation}
i.e., instead of static $\phi_i$ and $\theta_j$ parameters, we now have evolving coefficients to estimate, which is in compliance with online learning. Let $\boldsymbol{\Phi}_t = [\phi_{t,1}, \phi_{t,2}, \ldots, \phi_{t,p}]^T$ and $\boldsymbol{\Theta}_t = [\theta_{t,1}, \theta_{t,2}, \ldots, \theta_{t,q}]^T$. Then, for the state vector
\begin{equation*}
    \boldsymbol{s}_{t, ARMA} := [\boldsymbol{\Phi}_t^T, \boldsymbol{\Theta}_t^T]^T \in \mathbb{R}^{p+q},
\end{equation*}
we introduce the state transition and the measurement equations as
\begin{align}\label{eq:arma_ss}
    \boldsymbol{s}_{t,ARMA} &= \boldsymbol{s}_{t-1,ARMA} + \boldsymbol{e}_t\\
    y_{t,ARMA} &= \boldsymbol{r}_{t}^T\boldsymbol{s}_{t,ARMA} + \varepsilon_t,
\end{align}
where $\boldsymbol{e}_t$ is an i.i.d. noise vector, which can be sampled from any distribution such as Gaussian, driving the evolution of the parameters and
\begin{equation*}
    \boldsymbol{r}_{t} := [y_{t-1,ARMA}, \ldots y_{t-p,ARMA},\,\varepsilon_{t-1},\ldots \varepsilon_{t-q}]^T.
\end{equation*}
We note that all components of the vector $\boldsymbol{r}_t$ are known as it comprises only the past data, i.e., at time $t$, for $i > 0$, all $y_{t-i,ARMA}$ are observed and $\varepsilon_{t-i}$ are unknown yet estimated as $\hat{\varepsilon}_{t-i} = y_{t-i,ARMA} - \hat{y}_{t-i,ARMA}$.

Now we extend this representation for a $\text{SARIMAX}(p,d,q)(P,D,Q,m)$ model described in \eqref{eq:sarimax_eq}, which can be done via augmenting $\boldsymbol{s}_{t,ARMA}$ and $\boldsymbol{r}_{t}$ vectors in \eqref{eq:arma_ss} with the seasonal and exogenous coefficients. Then, we have
\begin{equation}
\begin{aligned}\label{eq:sarimax_ss}
    \tilde{\boldsymbol{\Phi}}_t &= [\phi_{t,1}, \ldots, \phi_{t,p}, \phi_{t,m}, \ldots, \phi_{t,mP}]^T\\
    \tilde{\boldsymbol{\Theta}}_t &= [\theta_{t,1}, \ldots, \theta_{t,q}, \theta_{t,m}, \ldots, \theta_{t,mQ}]^T\\
    \boldsymbol{s}_{t,SX} :&= [\tilde{\boldsymbol{\Phi}}_t^T, \tilde{\boldsymbol{\Theta}}_t^T, \boldsymbol{\beta}_t^T]^T\\
    \tilde{\boldsymbol{r}}_t :&= [y_{t-1,SX}, \ldots ,y_{t-p,SX},\\
                                  &\,\,\,\,\,\,\,\,\,y_{t-m,SX}, \ldots ,y_{t-mP,SX},\\
                                  &\,\,\,\,\,\,\,\,\,\varepsilon_{t-1}, \ldots, \varepsilon_{t-q},\varepsilon_{t-m},\ldots,\varepsilon_{t-mQ},\\
                                  &\,\,\,\,\,\,\,\,\,{x}_{t,1}, \ldots, {x}_{t,d}]^T\\
    \boldsymbol{s}_{t,SX} &= \boldsymbol{s}_{t-1,SX} + \boldsymbol{e}_t\\
    y_{t,SX} &= \tilde{\boldsymbol{r}}_{t}^T\boldsymbol{s}_{t,SX} + \varepsilon_t.
\end{aligned}
\end{equation}
In \eqref{eq:sarimax_ss}, the size of the state $\boldsymbol{s}_{t,SX}$ is $p+q+P+Q+d$ which is exactly the number of parameters to estimate in a SARIMAX model. We note that we assume the sequence $y_t$ is already ordinarily and/or seasonally differenced so that we do not account for differencing in the state space formulation, as in the Hamiltonian representation \cite{ss_hamilton}, and differences will be undone in prediction time, if any.

Given \eqref{eq:lstm_state_transition}, \eqref{eq:lstm_measurement} and \eqref{eq:sarimax_ss}, we can now concatenate the state vectors of an LSTM network and a SARIMAX model as
\begin{equation}\label{eq:joined_state}
    \begin{aligned}
        \boldsymbol{s}_t &:= \begin{bmatrix}
        \boldsymbol{s}_{t, LSTM}\\
        \boldsymbol{s}_{t, SX}\\
        \end{bmatrix}\\
                         &= [\boldsymbol{c}_t^T, \boldsymbol{h}_t^T, \boldsymbol{\theta}_t^T,
                            \tilde{\boldsymbol{\Phi}}_t^T, \tilde{\boldsymbol{\Theta}}_t^T, 
                             \boldsymbol{\beta}_t^T
                             ]^T
    \end{aligned}
\end{equation}
and arrive at the hybrid state space model as
\begin{equation}\label{eq:joined_state_space}
    \begin{aligned}
        \boldsymbol{s}_t &= \Omega(\boldsymbol{s}_{t-1}) + \boldsymbol{\eta}_t\\
        y_t &= y_{t,LSTM} + y_{t,SX} + \widetilde{\varepsilon_t}\\
            &= \boldsymbol{w}_t^T \boldsymbol{h}_t + \tilde{\boldsymbol{r}}_{t}^T\boldsymbol{s}_{t,SX} + \widetilde{\varepsilon_t}\\
            &= \boldsymbol{\lambda}_t^T\boldsymbol{s}_t + \widetilde{\varepsilon_t},
    \end{aligned}
\end{equation}
where $\widetilde{\varepsilon_t}$ and $\boldsymbol{\eta}_t$ are each i.i.d. noise processes that can be sampled from any distribution such as Gaussian, $\Omega(.)$ is a nonlinear function that applies the nonlinearities depicted in \eqref{eq:lstm_state_transition} to LSTM's internal state vectors $\boldsymbol{c}_t$ and $\boldsymbol{h}_t$ in $\boldsymbol{s}_t$, and otherwise behaves as a shifting operator, e.g., yields $\tilde{\boldsymbol{\Phi}}_{t - 1}$ when subjected to $\tilde{\boldsymbol{\Phi}}_{t}$, and $\boldsymbol{\lambda}_t$ is the gathered state coefficient vector comprising of zeros corresponding to $\boldsymbol{c}_t$ and $\boldsymbol{\theta}_t$ and otherwise carries $\boldsymbol{w}_t$ and $\tilde{\boldsymbol{r}}_{t}$. As for the measurement equation, we directly sum up the base models' predictions for the final output. We note that $\widetilde{\varepsilon_t}$ is assumed to be independent of each component of $\boldsymbol{\eta}_t$ for each time instant $t$.

In the next section, we present a particle filtering-based solution for the nonlinear system in \eqref{eq:joined_state_space}.

\subsection{Estimation with Particle Filtering}
Here, we introduce a particle filtering-based algorithm to find the unknown parameters in \eqref{eq:joined_state_space}. Our main aim is to estimate the recursively defined state vector $\boldsymbol{s}_t$ at each time step in an online manner after having observed measurements $\{\boldsymbol{y}_t\}$ by then. In a probabilistic point of view, this translates to obtaining the conditional distribution $p(\boldsymbol{s}_t\,|\,\boldsymbol{y}_{0:t})$, where $\boldsymbol{y}_{0:t}$ represents a slice of the vector $\boldsymbol{y}$ from its $0$\textsuperscript{th} component to $t$\textsuperscript{th} components, which is the set of observed measurements given as $\boldsymbol{y}_{0:t}$ = $\{y_0,y_1,\dots,y_t\}$.

As the state vector is recursively evolving, a recursive formulation for this distribution can be found by noting that the conditional joint distribution of $\boldsymbol{s}_{0:t}$ = $\{\boldsymbol{s}_0, \ldots, \boldsymbol{s}_t\}$ can be factored as
\begin{equation}\label{eq:prob_1}
    p(\boldsymbol{s}_{0:t}\,|\,\boldsymbol{y}_{0:t}) = Z\,p(\boldsymbol{s}_0\,|\,\boldsymbol{y}_0) \prod_{i = 1}^t p(\boldsymbol{y}_{i}\,|\,\boldsymbol{s}_{i}) p(\boldsymbol{s}_{i}\,|\,\boldsymbol{s}_{i-1}),
\end{equation}
where $Z$ is a constant normalizing factor. Using Bayes' theorem and with some algebra, we arrive from \eqref{eq:prob_1} to the recursive relation
\begin{equation}\label{eq:prob_2}
    p(\boldsymbol{s}_{0:t}\,|\,\boldsymbol{y}_{0:t}) = \frac{p(\boldsymbol{y}_{t}\,|\,\boldsymbol{s}_{t})\,p(\boldsymbol{s}_{t}\,|\,\boldsymbol{s}_{t-1})}{p(\boldsymbol{y}_{t}\,|\,\boldsymbol{y}_{0:t-1})}p(\boldsymbol{s}_{0:t-1}\,|\,\boldsymbol{y}_{0:t-1}),
\end{equation}
i.e., we can reach the desired distribution $p(\boldsymbol{s}_{0:t}\,|\,\boldsymbol{y}_{0:t})$ recursively via $p(\boldsymbol{s}_{0:t-1}\,|\,\boldsymbol{y}_{0:t-1})$. However, due to analytical intractabilities, e.g., the right hand side in \eqref{eq:prob_2} requiring analytically unsolvable integrals, we proceed with estimating these probabilities.

To this end, we apply the particle filtering to model the distribution containing the state vector in \eqref{eq:prob_2} via a discrete random measure \cite{PF_theory_1}, i.e.,
\begin{equation}\label{eq:pf_discrete}
    p(\boldsymbol{s}) \approx \sum_{i=1}^N \omega ^{(i)} \delta(\boldsymbol{s} - \boldsymbol{s}^{(i)}),
\end{equation}
where $\delta(.)$ is the Dirac's impulse function, $\boldsymbol{s}^{(i)}$ are the \emph{particles} simulating the state, $\omega ^{(i)}$ are the \emph{weights} associated with each particle and $N$ is the total number of particles. Formally, a discrete random measure $\mathcal{S} := \{\boldsymbol{s}^{(i)}, \omega ^{(i)}\}$ for $i \in \{1, \ldots, N\}$ provides the random particle-weight pairs. With this notion, the desired optimal state estimate (in the mean square error sense) is given as
\begin{equation}\label{eq:discretized_exp}
\begin{aligned}
    \mathbb{E}[\boldsymbol{s}_t\,|\,\boldsymbol{y}_{0:t}] &\approx \int \boldsymbol{s}_t\,\sum_{i=1}^N \omega_t^{(i)} \delta(\boldsymbol{s}_t - \boldsymbol{s}_t^{(i)})\,d\boldsymbol{s}_t\\
    &= \sum_{i=1}^N \omega_t^{(i)}\boldsymbol{s}_t^{(i)},
\end{aligned}
\end{equation}
i.e., a weighted sum of the particles. As $N$ goes to infinity, the equality is exact \cite{PF_theory_2}. The random measure $\mathcal{S}$ providing the particles is tied to the true conditional distribution of the state vector; however, it is in general unsuitable to sample from that distribution as it is unknown. Therefore, an easier-to-sample distribution which is related to the true distribution up to a proportionality is introduced, which is the importance function \cite{PF_theory_1}. Hence, using the following equivalence
\begin{equation}
    \begin{aligned}
    \mathbb{E}_p[f(x)] &= E_q\bigg[\frac{p(x)}{q(x)}f(x)\bigg]\\
    &\approx \frac{1}{N}\sum_{i=1}^N \frac{p(x_i)}{q(x_i)} f(x_i),
    \end{aligned}
\end{equation}
where $\mathbb{E}_r$ denotes the expectation with respect to the distribution $r(.)$, we reach
\begin{equation}\label{eq:p_to_q_with_state}
    \begin{aligned}
        \mathbb{E}_p[\boldsymbol{s}_t\,|\,\boldsymbol{y}_{0:t}] &= \mathbb{E}_q\bigg[\boldsymbol{s}_t\,\frac{p(\boldsymbol{s}_t\,|\,\boldsymbol{y}_{0:t})}{q(\boldsymbol{s}_t\,|\,\boldsymbol{y}_{0:t})}\,\bigg|\,\boldsymbol{y}_{0:t}\bigg]\\
    &\approx \frac{1}{N}\sum_{i=1}^N \frac{p(\boldsymbol{s}_t^{(i)}\,|\,\boldsymbol{y}_{0:t})}{q(\boldsymbol{s}_t^{(i)}\,|\,\boldsymbol{y}_{0:t})}\,\boldsymbol{s}_t^{(i)}.
\end{aligned}
\end{equation}
Then, it follows from \eqref{eq:discretized_exp} and \eqref{eq:p_to_q_with_state} that, the (unnormalized) weights are
\begin{equation}\label{eq:unnormalized_weights}
    \omega_t^{*(i)} = \frac{p(\boldsymbol{s}_t^{(i)}\,|\,\boldsymbol{y}_{0:t})}{q(\boldsymbol{s}_t^{(i)}\,|\,\boldsymbol{y}_{0:t})},
\end{equation}
from which the particle weights of $\mathcal{S}$ are given as
\begin{equation}
    \omega_t^{(i)} = \frac{\omega_t^{*(i)}}{\sum_{i=1}^N \omega_t^{*(i)}}.
\end{equation}
Since the weights are time-varying in the online learning framework, we derive a transition equation for the weights as well. This again relies on the recursive nature of the problem, by which we arrive at \cite{PF_theory_2}
\begin{equation}\label{eq:weights_upd_1}
    \omega_t^{(i)} = \frac{p(y_t|\boldsymbol{s}_t^{(i)})\,p(\boldsymbol{s}_{t}^{(i)}|\boldsymbol{s}_{t-1}^{(i)})}{q(\boldsymbol{s}_t^{(i)}|\boldsymbol{s}_{t-1}^{(i)}, y_t)}\omega_{t-1}^{(i)}.
\end{equation}
We choose the importance distribution $q(.)$ in \eqref{eq:weights_upd_1} to be the \emph{prior} importance function as commonly done in the literature to minimize the variance of the weights over time with easier computational conditions \cite{PF_theory_1}; hence, we have $q(\boldsymbol{s}_t^{(i)}\,|\,\boldsymbol{s}_{t-1}^{(i), y_t}) := p(\boldsymbol{s}_t^{(i)}\,|\,\boldsymbol{s}_{t-1}^{(i)})$.

Then, \eqref{eq:weights_upd_1} simplifies to give the transition of the weights as
\begin{equation}\label{eq:weight_upd_2}
    \omega_t^{(i)} = p(y_t|\boldsymbol{s}_t^{(i)})\,\omega_{t-1}^{(i)}.
\end{equation}
Finally, to deal with the well-known degeneracy problem \cite{PF_theory_1, PF_theory_2}, where the variance of the weights becomes too high to skew the weight distribution to an impulsive one, i.e., vast majority of the particles converge to nearly zero weights. This leads to undeserved attention to those weights in updating them via \eqref{eq:weight_upd_2}. This issue is addressed with the \emph{effective} sample size \cite{PF_N_eff} given as 

\begin{equation*}
    N_\text{eff} := \frac{1}{\sum_{i=1}^N {\omega_t^{(i)}}^2},
\end{equation*}
which measures the degeneracy such that if it is very low, the variance of the weights are very high; in which case, we can eliminate the aforementioned particles with very small weights and instead explore the large weighted particles by resampling \cite{PF_theory_1}.

The overall online prediction algorithm is outlined in \textbf{Algorithm \ref{alg:pf}}. For each particle, we update the candidate state vectors with the prior importance function, i.e., using the state transition equations. The corresponding weights evolve via multiplication with the likelihood of the associated state vector for producing the ground truth. Upon updates, we normalize weights mainly to resemble a probability distribution to be used in resampling, which happens when the effective size $N_\text{eff}$, a measure of variance of weights, goes below a certain threshold, e.g., half the original population $N$. Resampling is choosing candidate state vectors out of particles with replacement according to the distribution that weights compose, after which the weights are uniformized for the next iteration. Finally, the state estimation is a simple weighted average of particles from \eqref{eq:discretized_exp} and we output the prediction using this estimate and \eqref{eq:joined_state_space}.

\textit{\textbf{Remark.}} Note that one can use other nonlinear state space estimation methods such as extended Kalman filtering (EKF) \cite{julier1997new}, unscented Kalman filtering (UKF) \cite{wan2000unscented}] instead of the partical filtering to solve \eqref{eq:joined_state_space}. However, we observe from our experiments that the partical filtering has provided superior performance, which is also common observation in other applications \cite{chopin2002sequential}, hence we derive and provide only the required equations for the particle filtering case. To use the EKF or UKF, one needs to only change \textbf{Algorithm \ref{alg:pf}} accordingly, i.e., our state space approach is generic. 

\begin{algorithm}
\caption{Particle Filtering based Online Learning}\label{alg:pf}
\begin{algorithmic}
\Require $N, \text{number of particles}$
\Require $N_T, \text{resampling threshold}$
\For{i = 1 to N}
\State $\text{Sample } \boldsymbol{s}_t^{(i)} \sim p(\boldsymbol{s}_t\,|\,\boldsymbol{s}_{t-1}^{(i)})$
\State $\omega_t^{(i)} \gets p(y_t|\boldsymbol{s}_t^{(i)})\,\omega_{t-1}^{(i)}$
\EndFor
\State Total weight: $Z = \sum_{i=1}^N \omega_t^{(i)}$
\State Normalization: $\omega_t^{(i)} \gets \omega_t^{(i)} / Z, \forall i$
\State Effective size: $N_{\text{eff}} \gets 1 / \sum_{i=1}^N {\omega_t^{(i)}}^2$
\If{$N_\text{eff} < N_T$}
    \State Resample $\boldsymbol{s}_t^{(i)}, \forall i$ as per \cite{PF_theory_1}
    \State Uniformize weights: $\omega_t^{(i)} \gets 1 / N, \forall i$
\EndIf
\State Compute the state estimate: $\hat{\boldsymbol{s}}_t \gets \sum_{i=1}^N \omega_t^{(i) }\boldsymbol{s}_t^{(i)}$
\State Compute the prediction: $\hat{y}_{t+1} \gets \boldsymbol{\lambda}_t^T \hat{\boldsymbol{s}}_t$
\end{algorithmic}
\end{algorithm}

\section{Experiments}\label{sec:experiments}
In this section, we illustrate the performance of the introduced architecture under different scenarios. Firstly, we consider the prediction/regression performance of the model over the well-known real life datasets such as Bank \cite{bank_dataset}, Elevators \cite{elev_puma_dataset}, Kinematics \cite{kine_dataset} and Pumadyn \cite{elev_puma_dataset} datasets in an online setting. We then perform simulations over the yearly, daily and hourly subsets of the M4 competition \cite{m4_comp} dataset, aiming one-step ahead forecasting. We lastly consider a financial dataset, Alcoa stock prices \cite{alcoa_dataset}, where we demonstrate the generic nature of our architecture by replacing the LSTM part of our model with a GRU network. In each experiment, the only preprocessing step we apply to the datasets is standardization, i.e., taking the mean off and dividing by the standard deviation. In offline settings (e.g., with the M4 dataset), we first extract the mean and the standard deviation of the features across the entire training split of the data, and scale both the training and testing data with those same statistics. In online settings (e.g., in financial datasets), we apply standardization cumulatively via storing a running mean and running standard deviation as the stream provides data. We note that we do not extract any additional features from the datasets; the LSTM network is tasked to automatically extract the features from the raw data in a nonlinear fashion. All the experiments were conducted using a computer with i7-6700HQ processor, 2.6 GHz CPU and 12 GB RAM. 

\subsection{Real Life Datasets}\label{sec:reallifedatasets}
In this section, we evaluate the performance of the introduced model in an online learning setting. To this end, we consider four real life datasets.

\begin{itemize}
	\item Bank dataset \cite{bank_dataset} is a simulation of queues in a series of banks using customers with varying level of complexity of tasks and patience. We aim to predict the fraction of customers that have left the queue without getting their tasks done. There are 8,192 samples and each sample has 32 features, i.e., $\boldsymbol{x}_t \in \mathbb{R}^{32}$.
	\item Elevators dataset \cite{elev_puma_dataset} is obtained from the experiments of controlling an F16 aircraft; the aim is to predict a numeric variable in range $[0.012, 0.078]$ related to an action on the elevator of the plane, given 18 features. The dataset has 16,599 samples.
	\item Kinematics dataset \cite{kine_dataset} consists of 8,192 samples that are a result of a realistic simulation of the dynamics of an 8-link all-revolute robot arm. We aim to predict the distance of the end-effector from a given target using 8 features.
	\item Pumadyn dataset \cite{elev_puma_dataset}, similar to Kinematics dataset, contains a realistic simulation of a Puma 560 robot arm. The goal is to predict the angular acceleration of one of the robot arm's links. There are 8,192 samples and 32 features.
\end{itemize}

We compare our model against five different models: the Naive model, SARIMAX (without a seasonal component), multilayer perceptron (MLP), light gradient boosting machine (LightGBM) and the conventional LSTM architecture. The Naive model predicts $\hat{y}_{t+1} = y_t$ for all time steps, i.e., it outputs the last measurement as its prediction for the next time step. For the SARIMAX model, we tune the number of AR and MA components using a stepwise algorithm outlined in \cite{auto_arima}. The MLP we use in our experiments approximate a nonlinear autoregressive process with exogenous regressors (ARX), i.e., the values of the input neurons are the lagged values of the target signal as well as any side information the data provides. We use a one-layer shallow network for which we tune the number of hidden neurons, learning rate and the activation function in the hidden layer (ReLU \cite{relu} is used for the output neuron). LightGBM \cite{lgb_paper} is a hard tree-based gradient boosting machine, which, with well-designed features, achieves effective sequential data prediction. We tune the number of trees, learning rate, bagging fraction, maximum number of leaves and regularization constants for the LightGBM. For the LSTM model, we tune the number of layers, number of hidden units in each layer and the learning rate. For our architecture, we search for the number of particles in the filter, the number of hidden units for the LSTM component, the ordinary and seasonal orders for the SARIMAX model from a random subset of a grid. Among these five models, however, only the Naive model, MLP and LSTM are amenable to online learning; for the other two models we proceed as follows. For SARIMAX, we use the underlying Kalman filter to perform training; for LightGBM, we could perform re-training over the accumulated data at each time step; however, since this would already become too much time consuming, we introduce a queue of fixed size, say 50, and when it is fully filled, we append the new data to the historical data to refit the model; the queue is then emptied, and so on.

To assess the performance of online prediction/regression, we measure the time accumulated mean squared errors (MSEs) of models, which is the MSE computed over an ever-expanding window, i.e.,
\begin{equation}\label{eq:cum_mse_formula}
    \text{CumMSE}({\boldsymbol{y}}_t, {\boldsymbol{\hat y}}_t) = \frac{\sum_{t'=1}^{t}(y_{t'} - \hat{y}_{t'})^2}{t},
\end{equation}
where $t$ denotes the time until which the cumulative error is computed, $\boldsymbol{y}_t$ and $\boldsymbol{\hat y}_t$ represent the ground truth and prediction vectors until and including time $t$. We normalize \eqref{eq:cum_mse_formula} with $t$ to provide smoother curves, i.e., to get the mean performance. 

In Table \ref{table:real_life}, we report the cumulative error at the last time step for all models. We do not apply any preprocessing to the data for the Naive, SARIMAX and LightGBM models since they do not need normalization, while applying the standardization described in Section \ref{sec:experiments} to the other models.
\begin{table}[!b]
    \centering
    \caption{Cumulative errors of the models on Bank, Elevators, Kinematics and Pumadyn datasets}
    \begin{tabular}{c c c c c} 
                   & Bank       & Elevators      & Kinematics      & Pumadyn  \\
        \hline
        Naive      & 0.0307     &  0.0714        &  0.1305         &  0.0218  \\   
        SARIMAX     & 0.0183     &  0.0465        &  0.0897         &  0.0107  \\
        MLP        & 0.0165     &  \textbf{0.0043}        & 0.0766 &  0.0019  \\
        LightGBM   & 0.0203     &  0.0234        &  0.0991         &  0.0143  \\
        LSTM       & 0.0190     &  0.0119        &  0.0884         &  0.0094  \\
        LSTM-SX & \textbf{0.0129} & 0.0098 & \textbf{0.0627} & \textbf{0.0013} \\
        \hline
    \end{tabular}
    \label{table:real_life}
\end{table}

The introduced model (LSTM-SX) performs better on all datasets except for Elevators over the final cumulative error metric compared to other models, where for this dataset the results are close to each other. For example, in the Pumadyn dataset where the number of exogenous variables is the highest, our model performs significantly better than all models except for MLP, for which the results are close given the scale. This suggests the importance of an automatic feature extractor from the raw sequential data. Furthermore, the hybrid architecture consistently achieves better scores than the pure LSTM network; this emphasizes the contribution of the SARIMAX part, which not only addresses time series specific components, e.g., linear effect of shocks, but also allows for a direct contribution of exogenous variables to the final prediction. Overall, the results demonstrate the significant effectiveness of the framework in an online setting.

\subsection{M4 Competition Datasets}\label{sec:m4competitiondatasets}
The M4 competition \cite{m4_comp} provided $100,000$ time series of varying length with yearly, quarterly, monthly, weekly, daily and hourly frequencies. In our experiments, we chose yearly, daily and hourly datasets. The yearly dataset consists of $23,000$ very short series with the average length being $31.32$ years. Therefore, inclusion of this dataset aims to assess the performance of the models under sparse data conditions. Next, the daily dataset has $4,227$ very long series where the average length is $2357.38$ days. This dataset helps assess how effective a model is in capturing long-term trends and time shifts. Lastly, the hourly dataset consists of $414$ series. The reason for including this subset is the dominant seasonal component. We experiment over all of the hourly series and randomly selected $500$ of daily and yearly series each. The prediction horizon for hourly, daily and yearly datasets are $48$, $14$ and $6$ time steps, respectively. We aim for one-step ahead predictions in an offline setting; we train and validate the models over the training split of the data, and then predict the test split of the data one step at a time \emph{without} refitting/updating the models with the arrival of new data. Even though our framework is aimed for online regression, it can be easily adapted for offline prediction via treating the static data as a repeated stream.

We compare our model with five models presented in Section \ref{sec:reallifedatasets}. For each dataset, we separate a validation sequence from the end of each time series as long as the desired prediction horizon. For example, a series in the hourly dataset has its last $48$ samples reserved for validation of the hyperparameters. The searched hyperparameters of models as well as the preprocessing setup are as described in Section \ref{sec:reallifedatasets}. We use the MAPE to evaluate the performance of the models across series, which is scale-independent and defined as
\begin{equation*}
    \text{MAPE}({\boldsymbol{y}}_H, {\boldsymbol{\hat y}}_H) = \frac{1}{H}\sum_{t=1}^{H}\, \abs{\frac{y_{H,t} - \hat{y}_{H,t}}{y_{H,t}}},
\end{equation*}
where $H$ is the prediction horizon (e.g., $48$ hours), ${\boldsymbol{y}}_H$ and ${\boldsymbol{\hat y}}_H$ denote the vector of true and predicted values over the horizon, respectively.

The mean MAPE scores of the models over the datasets are presented in Table \ref{table:m4}. The LSTM-SX model achieves better performance than other models in all datasets, albeit with a close runner up in the daily dataset, LightGBM. In the yearly dataset, which consists of sparse data, LSTM-SX model adapts quickly than other models, especially the LSTM network. This implies the linear SARIMAX part helps attenuate the overfitting nature of a complex network, i.e., providing a regularization effect. On the other hand, in the daily dataset, which has very long sequences, our model also outperforms the others, suggesting that the history-caring feature extractor LSTM part in the architecture is in use as well. Lastly, our model's better score in the highly seasonal hourly dataset shows that the time series specific component, SARIMAX, models the seasonality well and justifies its presence in the integrated architecture. Collectively, the introduced model achieves better on average over several M4 competition datasets of varying characteristics.
\begin{table}[!t]
    \centering
    \caption{Mean MAPE performances of the models across M4 datasets}
    \begin{tabular}{c c c c} 
                   &         Hourly   &          Daily   &          Yearly \\
        \hline
        Naive      &         0.50915  &         0.09608  &         0.67152  \\ 
        SARIMAX    &         0.15549  &         0.05425  &         0.20032  \\ 
        MLP        &         0.19380  &         0.02940  &         0.22447  \\ 
        LightGBM   &         0.19468  &         0.01602 &         0.38881\\ 
        LSTM       &         0.20282  &         0.06448  &         0.18032  \\ 
        LSTM-SX & \textbf{0.13761} &         \textbf{0.01575} &   \textbf{0.14120} \\
        \hline
    \end{tabular}
    \label{table:m4}
\end{table}

\subsection{Ablation Study}\label{sec:financialdatasets}
Here, we show the generic nature of our architecture via substituting the LSTM network with a gated recurrent unit (GRU) network \cite{gru_paper}, which is another powerful RNN variant. A GRU cell, much like an LSTM cell, is geared with gated routing mechanisms to avoid the vanishing gradient problem \cite{rnn_vanishes}. Its transition equations analogous to \eqref{eq:lstm_internals} are given as
\begin{equation}\label{eq:gruforward}
\begin{aligned}
    \boldsymbol{z}_t &= \sigma(\boldsymbol{W}_z \,\{\boldsymbol h_{t-1}, \boldsymbol x_t\} + \boldsymbol b_z)\\
    \boldsymbol{q}_t &= \sigma(\boldsymbol{W}_q \,\{\boldsymbol h_{t-1}, \boldsymbol x_t\} + \boldsymbol b_r)\\
    \tilde{\boldsymbol h}_t &= \tanh(\boldsymbol{W}_h \,\{\boldsymbol{q}_t \odot \boldsymbol h_{t-1}, \boldsymbol x_t\} + \boldsymbol b_h)\\
    \boldsymbol{h}_t &= (1 - \boldsymbol{z}_t) \odot \boldsymbol{h}_{t-1} + \boldsymbol{z}_t \odot \tilde{\boldsymbol h}_t,
\end{aligned}
\end{equation}
where $\boldsymbol h_t$ is the hidden state vector, $\boldsymbol x_t$ is the input, $\boldsymbol{z}_t$ and $\boldsymbol{q}_t$ are the update and reset gate vectors, respectively, and $\tilde{\boldsymbol h}_t$ is the candidate state vector, whose corresponding stacked weight matrices are $\boldsymbol W_z, \boldsymbol W_q$ and $\boldsymbol W_h$. The corresponding biases are represented by $\boldsymbol b_z, \boldsymbol b_r$ and $\boldsymbol b_h$. To implement a GRU-SX, we directly replace the LSTM equations with GRU equations; this shows the flexibility of the framework.
\begin{figure*}[!h]
\centering
\includegraphics[width=0.8\linewidth]{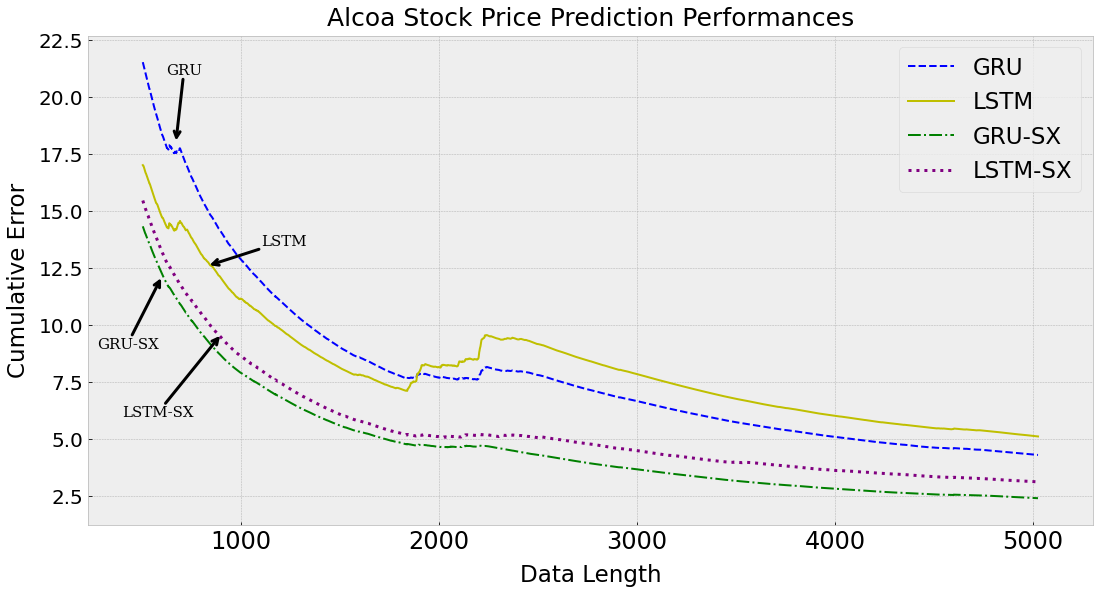}
\caption{Time accumulated errors of standalone and hybrid models for the Alcoa stock price dataset for the period 2000-2020.}
\label{fig:alcoa_perf}
\end{figure*}
We consider four models in this section: LSTM, GRU, LSTM-SX and GRU-SX. We assess the performances of them against a financial dataset in an online setting. We use the Alcoa stock price dataset \cite{alcoa_dataset}, which is comprised of the daily stock price values between the years 2000 and 2020. As side information, we use the lagged prices with a window size of 5. We only apply standardization to the price values in a cumulative manner. We hold out $10\%$ of the data from the beginning and use it for the hyperparameter validation. Tuned hyperparameters are the same as those in M4 datasets. Fig. \ref{fig:alcoa_perf} illustrates the performance of the models as the data length varies where we plot the cumulative MSEs over time via \eqref{eq:cum_mse_formula}. 

We observe that the introduced hybrid models consistently achieve lower accumulated errors than the purely recurrent models, showing the efficacy of the framework. GRU network begins with a rather big error margin but closes the gap after ~2000 data points and performs better than LSTM then on. The hybrid models perform close to each other yet the GRU-SX model performs consistently better. The standalone LSTM and GRU models are not only rather slower to react to the sudden change in the stock prices around 2,200 data points, which corresponds to the global financial crisis in the late 2008, but also face a noticeable increase on cumulative error unlike the hybrid models.

\section{Conclusion}\label{sec:conclusion}
We studied nonlinear prediction/regression in an online setting and introduced a hybrid/ensemble architecture composed of an LSTM and a SARIMAX model. Unlike the current practice of disjoint ensembles, we unify an enhanced RNN, acting as a feature extractor from the raw sequential data, and the SARIMAX model, addressing time series intricacies robustly, e.g., seasonality, in a joint manner with a single state space. We thereby remove the need for domain specific feature engineering and achieve an efficient mix of nonlinear and linear components, while enjoying joint optimization for the integrated architecture thanks to the novel state space formulations introduced. We train the hybrid model via particle filtering and present the associated update equations for the learnable parameters. We emphasize that our architecture is generic so that one can use other recurrent deep learning architectures for feature extraction such as RNNs and GRUs, other traditional models such as ETS for effective time series modelling and other state space training approaches such as EKF, UKF. We achieve significant and consistent performance gains in the experiments over conventional methods on various well-known benchmark and competition datasets. We also provide the source code for our model for further research and reproducibility of our results. 


\bibliographystyle{IEEEtran}
\bibliography{refs}

\end{document}